\documentclass[aoas]{imsart}
\usepackage{graphicx}

\usepackage{algorithmic}
\usepackage{algorithm}

\RequirePackage{amsthm,amsmath,amsfonts,amssymb}
\RequirePackage[authoryear]{natbib}

\startlocaldefs
\endlocaldefs

\begin{document}

\begin{frontmatter}
\title{Emergent Granger Causality in Neural Networks: Can Prediction Alone Reveal Structure?}
%\title{A sample article title with some additional note\thanksref{T1}}
\runtitle{Can Prediction Alone Reveal Structure?}
%\thankstext{T1}{A sample of additional note to the title.}

\begin{aug}
%%%%%%%%%%%%%%%%%%%%%%%%%%%%%%%%%%%%%%%%%%%%%%%
%% Only one address is permitted per author. %%
%% Only division, organization and e-mail is %%
%% included in the address.                  %%
%% Additional information such as            %%
%% identifying the corresponding author must %%
%% be included in in the Acknowledgments     %%
%% section if necessary.                     %%
\author[A]{\fnms{Malik Shahid}~\snm{Sultan}\ead[label=e1]{malikshahid.sultan@kaust.edu.sa}},
\author[B]{\fnms{Hernando}~\snm{Ombao}\ead[label=e2]{hernando.ombao@kaust.edu.sa}}
\and
\author[B]{\fnms{Maurizio}~\snm{Filippone}\ead[label=e3]{maurizio.filippone@kaust.edu.sa}}

%%%%%%%%%%%%%%%%%%%%%%%%%%%%%%%%%%%%%%%%%%%%%%
%% Addresses                                %%
%%%%%%%%%%%%%%%%%%%%%%%%%%%%%%%%%%%%%%%%%%%%%%
\address[A]{BESE Division,
King Abdullah University of Science and Technology (KAUST)%
\printead[presep={,\ }]{e1}}

\address[B]{CEMSE Division,
King Abdullah University of Science and Technology (KAUST)%
\printead[presep={,\ }]{e2,e3}}
\end{aug}

\begin{abstract}
Granger Causality (GC) offers an elegant statistical framework to study the association between multivariate time series data. Vector autoregressive models (VAR) are simple and easy to fit, but have limited practical application because of their inherent inability to capture more complex (e.g., non-linear) associations. Numerous attempts have already been made in the literature that exploit the functional approximation power of deep neural networks (DNNs) for GC. However, these methods treat GC as a variable selection problem. We present a novel paradigm for investigating the learned GC from a single neural network used for joint modeling of all components of multivariate time series data, which is essentially linked with prediction and assessing the distribution shift in residuals. A deep learning model, with proper regularization, may learn the true GC structure when jointly used for all components of the time series when there is sufficient training data. We propose to uncover the learned GC structure by comparing the model uncertainty or distribution of the residuals when the past of everything is used as compared to the one where a specific time series component is dropped from the model. We also compare the effect of input layer dropout on the ability of a neural network to learn GC. We show that a well-regularized model can learn the true GC structure from the data without  explicitly adding terms in the loss function that guide the model to select variables or perform sparse regression under specific settings. We also provide a comparison of deep learning architectures such as CNN, LSTM and transformer models on their ability to discover Granger Causality. The numerical experiments demonstrate that, compared to sparse regression models, a simple joint model is a strong baseline for learning the true GC which has the advantage that it does not require tuning of many extra hyper-parameters.

\end{abstract}

\begin{keyword}
\kwd{Machine Learning}
\kwd{Neural Networks}
\kwd{Granger Causality}
\kwd{Deep Learning}
\kwd{Time Series}
\end{keyword}

\end{frontmatter}

\section{Introduction}
Neural Networks are popular models to learn parametric functions, useful for various Machine Learning (ML) tasks, from data. Deep Neural Networks (DNNs) excel at predictive modeling tasks \citep{tercan2022machine,ismail2019deep}, and recent advances in Natural Language Processing (NLP) \citep{yang2025recent} and Computer Vision (CV) \citep{zhang2022survey} demonstrate tremendous predictive and representation learning power of these models.
\par
Multivariate time series data consists of realizations of temporal stochastic processes (e.g., brain signals recorded from many channels). Understanding the connectivity structure in the data is important in many applications. For instance, in neuroscience, it is important to understand the functional and effective connectivity between different brain regions, in finance it is important to discover the lead-lag relationships between stocks, and in biology it is of great interest to uncover the gene-regulatory path ways. We consider analyzing multi-channel brain signals 
%Some of the primary goals in analyzing multivariate time series data are  
with the goal of predictive modeling and uncovering the cross-dependence structure between components of the brain network. One potential impact of this work is that physiological biomarkers (including causal structure in brain networks) have the potential to differentiate between patient subgroups \citep{stiglic2020interpretability, abdullah2021review}.
\par
There are existing time series tools (e.g., temporal and spectral)
that take into account specific temporal or spectral information in time series data to 
that quantify the dependence structure in a network, including cross-correlation, mutual information, coherence, and spectral transfer entropy \citep{ombao2024spectral, redondo2023measuring}.
Granger Causality (GC) provides another framework for quantifying dependence in multichannel brain signals. GC essentially measures the ability of the past of one channel (indexed by $i$) to predict the future of another channel (indexed by $j$). This framework, proposed in \citet{granger1969investigating},
relies on two main assumptions:
\begin{enumerate}
    \item Cause happens prior to its effect
    \item Cause has unique information about the future of its effect
\end{enumerate}

More formally, suppose that ${\bf X}(t) = [X_1(t), \ldots, X_P(t)]$ is a $P$-dimensional real-valued time series at time $t$. To test if time series $X_i$ is a Granger cause of time series $X_j$, we consider two models for predicting $X_j$: the first uses all past data to predict the future of time series $X_j$ (hence referred to as the {\em full model}); the second model uses all past data {\it except} for time series $X_i$, (hence called the {\em reduced model}. The two models are fit to the observed data, and residuals (or errors) are conpared for the full and reduced models. This is rigorously conducted by formulating a test statistic based on the discrepancy between the fit of the two models. The observed test statistic is compared to a reference distribution under the null hypothesis of no Granger causality from $X_i$ to $X_j$. 

Linear and non-linear models have been proposed in the literature to estimate GC from the data. Among linear models, Vector autoregressive (VAR) models are most widely employed for the purpose of exploring and testing GC. A more recent alternative is deep learning (DL) which is being extensively explored for GC formulated under sparse regression \citep{tank2021neural, cheng2023cuts, khanna2019economy, marcinkevivcs2021interpretable}. One commonly used framework,  introduced in \citet{tank2021neural}, builds a case for the use of separate neural networks for the modeling of individual time series components ${\bf X}(t)$. Sparsity-inducing penalties are applied on the input layer of these component-wise models to learn the dependence structure from the data. The weights of the input layer are gradually shrunk to zero using proximal gradient descent. GC is then estimated from the weights of the input layer of these component-wise models. A component wise NN for predicting the component $X_p$ is denoted  $f_{p,\theta}$ with $g_{p,\omega}$ as the associated input layer, models the dynamics of the time series $X_p$ using the past $K$ lags of the multivariate time series ${\bf X}(t)$. The parameter vector $\theta$ stores the learnable parameters (weights and biases) of the hidden layers while $\omega$ contains the learnable parameters for the input layer.
\begin{equation}
    X_p(t) = f_{p,\theta}(g_{p,\omega}({\bf X}(t-1:t-K))) + e(p,t).    
\end{equation}
Group Lasso is applied on $g_{p,\omega}$ to sparsify the input layer weights and GC is estimated from the weights of $g_{p,\omega}$. As one may think this can become quickly computationally expensive if the number of time series grows or if the number of datasets grow. One such example is in healthcare where one may be interested in learning the GC between EEG recording of each subject separately for the purpose of personalized medicine or group level analysis.

Authors in \citet{tank2021neural} argue that inference on the weights of the neural network cannot lead to meaningful conclusions for estimation of GC when multivariate time series data is modeled using a single auto-regressive model, because of the weight sharing between all time series. However, in our work, we present a different point of view for GC. We agree on the aforementioned statement that if a joint model is trained it is not straightforward to estimate GC from the weights of the network. However, GC is linked to prediction, and if a joint model is trained to predict the future dynamics of a multivariate time series data, it may learn some GC structure which has to be uncovered. Also for the purpose of interpretation it becomes essential to uncover the learned GC structure for any black-box model like neural networks. This leads towards the motivation of the present work, where we show how to uncover the GC structure learned by the joint model, and how effective just predictive modeling is to learn the true Granger causal structure without sparse regression. In our work, we show that even if a joint model is being used to model the multivariate time series data, post-hoc model explanations about GC can still be performed, by comparing the distribution of the residuals from the fit of the joint model under specific input layer configurations. In this work, we make the following key contributions:\begin{enumerate}
    \item \textbf{Neural Granger Causality Modeling with Monte Carlo (MC) Dropout :} We investigate whether neural networks jointly used to model multivariate time series trained with MC Dropout can learn the underlying Granger causal structure.

    \item \textbf{Post-hoc GC Extraction without Sparse Priors:} We propose a novel approach to uncover the learned GC from a trained neural networks without relying on sparse regression techniques or introducing additional tunable hyper-parameters.

    \item \textbf{Causality Quantification Mechanism:} We develop a mechanism for quantifying Granger causal influence in settings where time series are modeled jointly by a neural network.

    \item \textbf{Role of Input Dropout:} We examine the impact of input layer dropout on the ability of neural networks to capture Granger causal relationships.

    \item \textbf{Architecture-wise Comparison:} We provide a comparative analysis of different neural architectures namely, Transformers~\citep{vaswani2017attention}, LSTMs~\citep{hochreiter1997long}, and CNNs~\citep{6795724} in terms of their capacity to recover Granger causality from time series data.
\end{enumerate} 

Where previous works proposed sparse regression-based methods, we simply try to uncover the learned GC by a trained model and study how well neural networks perform structure learning while just trained on the task of regression. We propose a method to uncover the learned GC from a single neural network that jointly models the multi-variate time series data, by comparing the distribution of the errors in the predictions of the model under general and reduced models. We also compare the effect of input layer dropout on the model's ability to learn GC from the data. To the best of our knowledge this is the first work in time series GC that explores the effect of input layer dropout during training and estimation of GC by comparing distribution of residuals.

\section{Related Work}
The problem of studying the dependence structure with GC has been extensively explored in statistics, ML, and DL. Traditionally, Vector Auto-Regressive (VAR) Models are used to infer GC from the multi-variate time series data. A linear VAR(K) model, models the future of the multivariate time series data $X(t)$ as a linear function of the past $K$ lags  \citep{lutkepohl2013introduction, shojaie2022granger}. To test whether a uni-variate time series indexed by $i$ is a Granger cause of a uni-variate time series indexed by $j$, full and restricted models are constructed and a test statistic is computed from the fit of these models. The test statistic is then compared with the null distribution of no Granger causality from $i$ to $j$. One of the primary limitations of these methods is their inability to model the non-linear associations in the data. Data transformations like kernel-based methods have been studied and proposed in the literature \citep{marinazzo2008kernel, sun2008assessing} to model the non-linear dynamics; however, the choice of these kernel functions has to be driven by domain knowledge. 
\par
ML and DL have dealt with the problem of GC estimation as a variable selection problem within sparse regression. Neural networks are becoming more and more popular because of their ability to learn data dependent transformations. Because of their ability to model complex associations in the data, these models are an excellent candidate for modeling time series data. However, because DNNs are essentially black-boxes, they excel at prediction tasks with limited explainability. This problem was identified and a solution of using component-wise models with sparse input layer was proposed in \citet{tank2021neural}. They proposed to use separate neural networks to model each time series separately. Constraints like group lasso and hierarchal group lasso that promote sparsity in the input layer of these component wise models are imposed. GC is then inferred from the weights of these input layers. Extending the same framework of \citet{tank2021neural}, \citet{khanna2019economy} made model adjustments and proposed the Economy Statistical Recurrent units (eSRU) model with gating to estimate GC. \citet{marcinkevivcs2021interpretable} proposed the Generalized VAR (GVAR) model, where the GC and the sign of the effect was estimated simultaneously; however, their model requires tuning of hyper-parameters including sparsity and smoothing parameters. Causal neural discovery from irregular high-dimensional time series was proposed in \cite{cheng2023cuts} and \cite{cheng2024cuts+} respectively. \citet{yin2022deep} proposes a framework to discover the GC from observational time series data in the presence of latent confounding.
\par
Greedy search for the best features or GC lag components was proposed by \citet{montalto2015neural}. Only lagged values of a time series component are added if they help improve the prediction of the future of another time series component. The framework of \citet{montalto2015neural} was extended in \citet{wang2018estimating}, where the base Multi-layer Perceptron (MLP) model was replaced with gated Recurrent Neural Networks (RNNs). 
\par
Previous works on neural Granger causality, lack the fundamental investigation on the ability of the neural networks to learn the GC from the time series data without sparse regression with joint modeling of multi-variate data. Also to the best of our knowledge its not evident how transformers, LSTMs or CNN differ in the ability to learn Granger Causality from time series data, under the task of prediction. Self-supervision training where the past of the few time series are dropped to predict the future of the time series, is also studied under input layer dropout where we study what happens to the models ability to learn Granger causal structure under this setting. 
\section{Method}

\begin{figure}[t]
\centering
\includegraphics[width=1\linewidth]{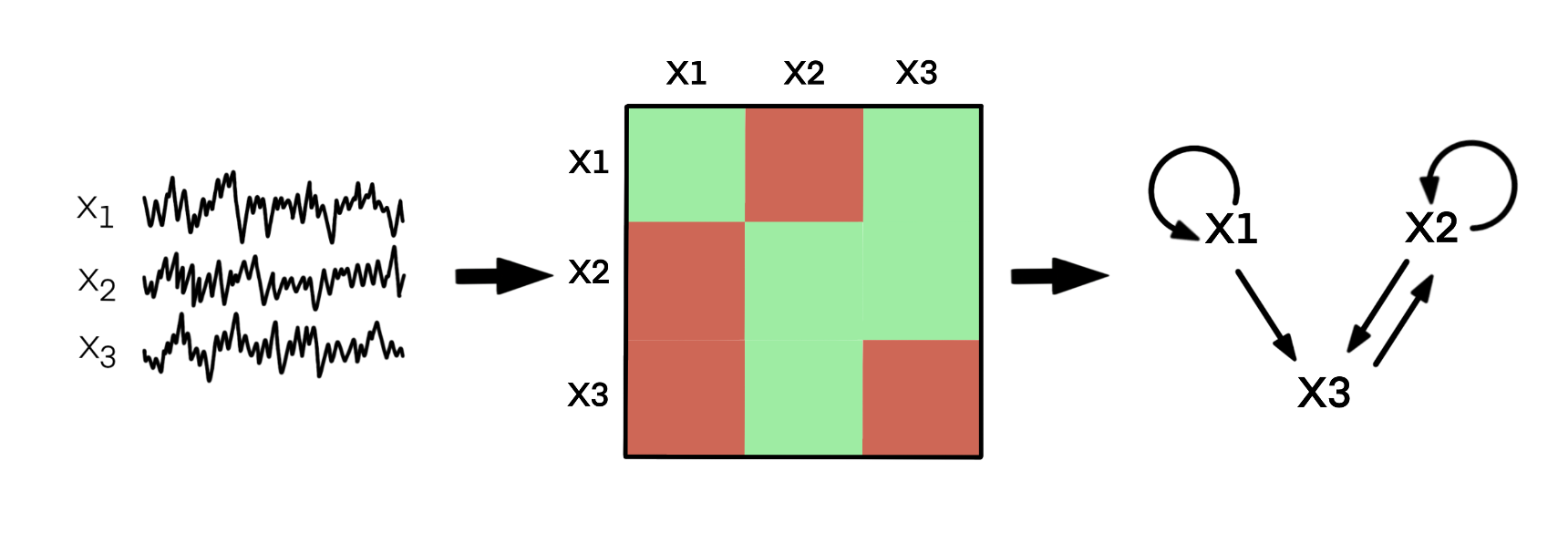}%
\vfill
\includegraphics[width=1\linewidth]{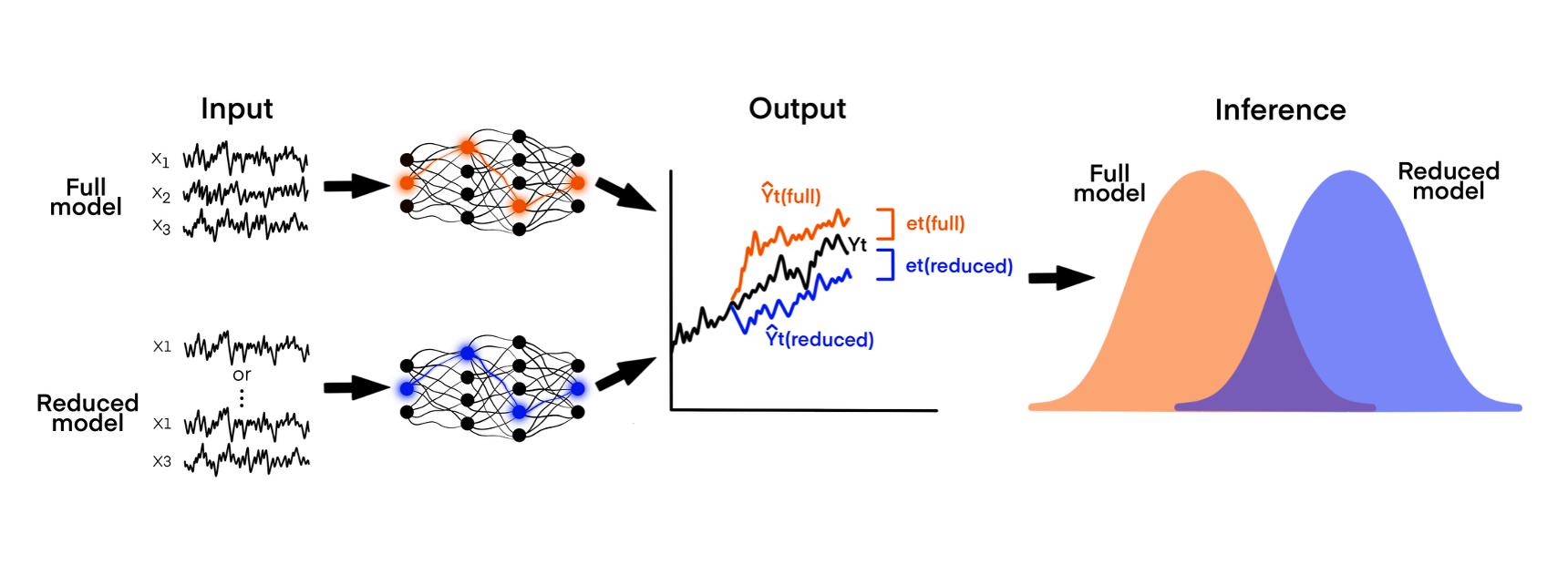}
\caption{Time series data is observed under some unknown Granger causal structure. A deep learning model $f_{\theta}$ is trained to predict the future dynamics of the time series data conditional on the past lags. Once a model $f_{\theta}$ has been trained and an estimate $\hat{\theta}$ has been obtained under dropout probability $\alpha^{*}$, multiple predictions under the test data can be made to obtain a distribution of the mean squared errors under the full and specific reduced model. These distributions can then be compared to quantify the probability of a time series Granger causing another time series. }
\label{fig:method}
\end{figure}

Let $X(t)$ $\in$ $R^P$ be the recording of a $P$ dimensional time series data at time point $t$. The goal is to estimate the $P \times P$ binary adjacency matrix, where if $j \rightarrow i$ then $A[i,j] =  1 $. Let $f_{\theta}$ represent the neural network used to model the dynamics of the system. $f$ is parameterized by $\theta$, which is a set of all weights and biases of the neural network. $\theta$ has to estimated from the data using backpropagation.
\par
In our experiments, we explore three settings:
\begin{enumerate}
    \item No input layer dropout, dropout is just incorporated in the hidden layers.
    \item Input layer and hidden layer dropout. Two forward passes are made through the model, one with the input layer dropout and one without input layer dropout.
    \item Input layer and hidden layer dropout. Only one forward pass is made through the model, with the input layer and hidden layer dropout being active.
\end{enumerate}

Let $m$ denote a P dimensional multi-hot encoded mask. The positions where 1 is present indicates that the past of these time series are being used by the model to predict the future. In the present setting all past lags of the time series $i$ are dropped if it has 0 in the mask at the index $i$. A general and reduced model along with forward pass inference, where no and few time series are dropped at random are given in equations \ref{eq:FM Dynamics}, \ref{eq:RM Dynamics}, \ref{eq:FM Pred} and \ref{eq:RM Pred}, respectively.
\begin{equation}
    X_{f}(t)  =  f_{\theta}(X(t-1:t-K)) + e_{f}(t)
    \label{eq:FM Dynamics}
\end{equation}

\begin{equation}
    X_{r}(t)  =  f_{\theta}(m * X(t-1:t-K)) + e_{r}(t)
    \label{eq:RM Dynamics}
\end{equation}

\begin{equation}
    \hat{X}_{f}(t)  =  f_{\hat{\theta}}(X(t-1:t-K))
    \label{eq:FM Pred}
\end{equation}

\begin{equation}
    \hat{X}_{r}(t)  =  f_{\hat{\theta}}(m * X(t-1:t-K))
    \label{eq:RM Pred}
\end{equation}

$X(t-1:t-K)$ refers to the past K lags of time series X, $*$ indicates element-wise product. $\hat{\theta}$ refers to the set of estimated weights and biases for the neural network using backpropagation of errors. If $f_{\theta}$, is trained in a fashion that it can be used for uncertainty quantification in the predictions, predictions under full and reduced models with specific masking structure incorporated in $m$ can be made along with uncertainty quantification in these predictions. Multiple passes can be made to create distributions of residuals under full and reduced models. The task at hand is then to compare the distributions under the full and reduced models for specific components of time series to check if one time series is a Granger cause for another one. This uncovers the model-dependent GC, which can be used to create explanations for the predictions.

\subsection{Training with Monte Carlo Dropout}
MC Dropout proposed in \cite{gal2016dropout} provides a computationally economical way for model uncertainty quantification in neural networks. Dropout is a famous regularization technique used in the training of neural networks. With MC dropout, dropout is active even in the prediction phase, yielding an ensemble of predictions that can be used for uncertainty quantification. The dropout probability can be tuned using cross-validation on multiple folds or with just one validation set. The loss functions for the models trained with MC dropout is given in equations \ref{eq:Loss Function for MC Dropout One Pass Full Model}, \ref{eq:Loss Function for MC Dropout Two passes}, and \ref{eq:Loss Function for MC Dropout One Pass Red Model} for three scenarios.

\begin{equation}
    \mathcal{L}(\theta, \alpha)_{\hbox{\scriptsize No ILD}}= \frac{1}{n} \sum_{i=1}^{n} \left( X(t) - \hat{X}_{f}(t) \right)^2 
    \label{eq:Loss Function for MC Dropout One Pass Full Model}
\end{equation}

where $\alpha$ is the dropout probability in the hidden layers which is a hyper-parameter and has to be tuned.
Equation \ref{eq:Loss Function for MC Dropout One Pass Full Model} refers to No Input Layer Dropout (No ILD) and is a mean squared error loss function for regression problems, in the present scenario it refers to the case where predictions are made from a model with just hidden layer dropout.
\begin{equation}
\begin{array}{rcl}
\mathcal{L}(\theta,\alpha)_{\hbox{\scriptsize DP ILD}} &=& \frac{1}{n} \sum_{i=1}^{n} \left( X(t) - \hat{X}_{f}(t) \right)^2 \\
&&+ \frac{1}{n} \sum_{i=1}^{n} \left( X(t) - \hat{X}_{r}(t) \right)^2
\end{array}
\label{eq:Loss Function for MC Dropout Two passes}
\end{equation}
 Equation \ref{eq:Loss Function for MC Dropout Two passes} refers to Dual Pass Input Layer Dropout (DP ILD), the overall loss has two components, the discrepancy in the prediction under no dropout on the input time series components and with random dropout. The dropout which we perform on the input layer is random. At any mini batch of data, uniformly 1 to $P-1$ time series can be dropped during forward pass. This ensures that in the worst case scenario there is at least the past of one time series to predict the future of everything when the dropout is active during the training. The intuition for training the model in this way is to study if by doing input layer dropout the model may learn full and multiple reduced models.

\begin{equation}
    \mathcal{L}(\theta,\alpha)_{\hbox{\scriptsize ILD}} = \frac{1}{n} \sum_{i=1}^{n} \left( X(t) - \hat{X}_{r}(t) \right)^2 
    \label{eq:Loss Function for MC Dropout One Pass Red Model}
\end{equation}
 Equation \ref{eq:Loss Function for MC Dropout One Pass Red Model} refers to Input Layer Dropout (ILD) training.
 We systematically investigate the effect of removing the forward pass with no dropout and only keep random dropout computations in the input layer. The idea behind this is to check the effect of only performing random dropout and the ability of the neural networks to learn Granger causal associations.

%\subsubsection{Training only with Input Layer Dropout}
%On clearly examining the equation \ref{eq:Loss Function for MC Dropout}, one can see that there are two forward passes through the model one with dropout and one without dropout. We also systematically investigate the effect of removing the forward pass with no dropout and only keep random dropout computations. The idea behind this is to check the effect of only performing random dropout and the ability of the neural networks to learn Granger causal associations. The loss function will then be reduced to the following form

%\begin{equation}
%    \mathcal{L}(\theta) = \frac{1}{n} \sum_{i=1}^{n} \left( X(t) - \hat{X}_{r}(t) \right)^2 + \lambda g(\theta).
%    \label{eq:Loss Function for MC Dropout one pass}
%\end{equation}

\subsection{Obtaining Distribution of Residuals}
Once $f_{\theta}$ has been trained, an estimate $\hat{\theta}$ has been obtained, and the model can be used to make predictions on unseen data. $Q$ forward passes can be made through the model to obtain $Q$ predictions while keeping the data constant. To decide if time series $i$ Granger causes times series $j$, $Q$ predictions for time series $j$ for the test set or holdout set are made with and without the past of time series $i$. We keep the same dropout structure in the hidden layers for both cases. $E_{j,f} = [e_{1,f},e_{2,f} \cdots e_{Q,f}]$ and $E_{j,i,r} = [e_{1,i,r},e_{2,i,r} \cdots e_{Q,i,r}]$ correspond to the distributions of the residuals under the full and reduced models (with and without the past of time series $i$) for the $j^{th}$ component of the time series.
\begin{algorithm}[t]
\caption{Obtaining Residual Distributions for Granger Causality: $i \rightarrow j$}
\label{alg:gc-residual-distributions}
\begin{algorithmic}[1]
\REQUIRE Trained neural network $f_{\hat{\theta}}$, dropout rate $\alpha^*$, test data $\{(x_{t-k}, x_t)\}_{t=1}^T$, number of forward passes $Q$, candidate cause index $i$, target index $j$
\ENSURE Residual sets $E_{j,f}$ (full model) and $E_{j,i,r}$ (reduced model)

\STATE Generate a fixed list of random seeds: $\mathcal{S} = \{s_1, s_2, ..., s_Q\}$

\STATE \textbf{Step 1: Full Model Predictions (all inputs)}
\FOR{$q = 1$ to $Q$}
    \STATE Set seed $s_q$
    \STATE Predict output: $\hat{x}_{t,j}^{(q)} = f_{\hat{\theta}}(x_{t-k})$
    \STATE Compute squared error between $\hat{x}_{t,j}^{(q)}$ and $x_{t,j}$
    \STATE Store this in $E_{j,f}$
\ENDFOR

\STATE \textbf{Step 2: Reduced Model Predictions (input $i$ masked)}
\FOR{$q = 1$ to $Q$}
    \STATE Set dropout seed $s_q$
    \STATE Mask series $i$: set $x'_{t-k} = m * x_{t-k}$, where $m[i] = 0$, others = 1
    \STATE Predict output: $\hat{x}_{t,j}^{(q)} = f_{\hat{\theta}}(x'_{t-k})$
    \STATE Compute squared error between $\hat{x}_{t,j}^{(q)}$ and $x_{t,j}$
    \STATE Store this in $E_{j,i,r}$
\ENDFOR

\RETURN Residual sets $E_{j,f}$ and $E_{j,i,r}$
\end{algorithmic}
\end{algorithm}

\subsection{Comparing Distributions of Residuals for GC}
\begin{algorithm}[t]
\caption{Detecting Granger Causality with Distribution Shift}
\label{alg:gc-classifier}
\begin{algorithmic}[1]
\REQUIRE Error sets $E_{j,f}$ (full model) and $E_{j,i,r}$ (reduced model), number of samples $M$
\ENSURE Granger causality probability $P(i \rightarrow j)$

\STATE Construct dataset $D = \{(e_m, y_m)\}_{m=1}^{2M}$, where $y_m = 1$ if $e_m \in E_{j,f}$ else $0$
\STATE Train binary classifier $\hat{g}$ on $D$ using cross-validation
\STATE For each $e_{m} \in E_{j,i,r}$, compute $p_m = \hat{g}(e_{m})$
\STATE Compute $P(i \rightarrow j) = \frac{1}{M} \sum_{m=1}^{M} (1 - p_m)$
\RETURN $P(i \rightarrow j)$
\end{algorithmic}
\end{algorithm}
After obtaining $E_{j,f}$ and $E_{j,i,r}$, the next step is to decide if $i$ Granger causes $j$ ($i \rightarrow j$). This can be done by comparing these distributions. One strategy is to look at the overlap of these distributions. The intuition is that conditional on the samples of the mean squared error from these distributions can we discriminate if these are coming from the full model or the reduced model? If the distributions do not overlap and if the mean of $E_{j,i,r}$ is greater then $E_{j,f}$ then it is an indication that removing the past of time series $i$ degrades the ability of the model to predict the future dynamics of time series $j$, therefore $i$ is a Granger cause of $j$ under this model. However if the distributions overlap significantly or if the mean of $E_{j,i,r}$ is less then $E_{j,f}$ then $i$ is not a Granger cause of $j$. The scenario where it happens that the mean of $E_{j,i,r}$ is less than mean of $E_{j,f}$, indicates that removing the time series $i$ leads towards a decrease in the model prediction error, which leads towards non-Granger causation.
\par
To classify if the samples are coming from a full or reduced model a binary classifier can be used. To check if two distributions are similar or overlap we can create a labeled data set $D = \{ e_{m}, y_{m}\}_{m=1}^{M}$ for $M$ training examples, where $y_{m}$ is binary label which is 1 if $e_{m} \in E_{j,f}$ else 0. A binary classifier can be trained on this dataset D using cross-validation. Suppose $\hat{g}$ is the classifier trained using cross validation. The probability that the samples from the reduced model are classified as of full model can be obtained using $\hat{g}$ to predict the  $P (y_{m} = 1 | e_m)$ $ \forall$ $ e_m \in E_{j,i,r}$. This also gives an extent to which $E_{j,i,r}$ overlaps with $E_{j,f}$. The average of this probability for all samples in $E_{j,i,r}$ can be used to quantify the degree of overlap.
\begin{equation}
    P(y_{m} = 1 | e_{m,r}) = \hat{g}_{p,q}(e_{m,r})
    \label{eq : DRE Trick}
\end{equation}
where $e_{m,r}$ is the $m^{th}$ sample from the distribution of mean squared errors under the reduced model. 
\par
The average of this probability for all samples in $E_{j,i,r}$ can be used to summarize this information. If both distributions can be clearly classified or separated by the classifier and if the mean of the distribution under the full model is less then that of the reduced model then we can conclude that $i \rightarrow j$. 
We can now quantify the probability of $i$ $\rightarrow$ $j$ as the mean of $1 - P(y_{m} = 1 | e_{m,r})$ $\forall$ $e_{m,r} \in E_{j,ir}$ as indicated in equation \ref{eq : DRE prob}.
  
\begin{equation}
    P(i \rightarrow j) = \frac{\sum_{m=1}^M 1 - P(y_{m} = 1 | e_{m,r})}{M}
    \label{eq : DRE prob}
\end{equation}

\section{Simulation Studies}
We start with a simple simulation study to analyze the ability of different  neural network base architectures Transformers, CNN and LSTM to uncover the true Granger causal structure from the observed time series data under the task of prediction. We generate time series data under fork, collider and chain structures with just 3 components. We also test the ability of unrestricted neural networks to identify the Granger causality for the VAR(2), Lorenz and Netsim benchmark simulations. We also compare the performance with the sparse regression based methods including cLSTM \citep{tank2021neural}, GVAR \citep{marcinkevivcs2021interpretable}, and CUTS \citep{cheng2023cuts}.
\subsection{Chain, Fork and Collider}
\begin{figure}[h]
    \centering
    \includegraphics[width=1.0\linewidth]{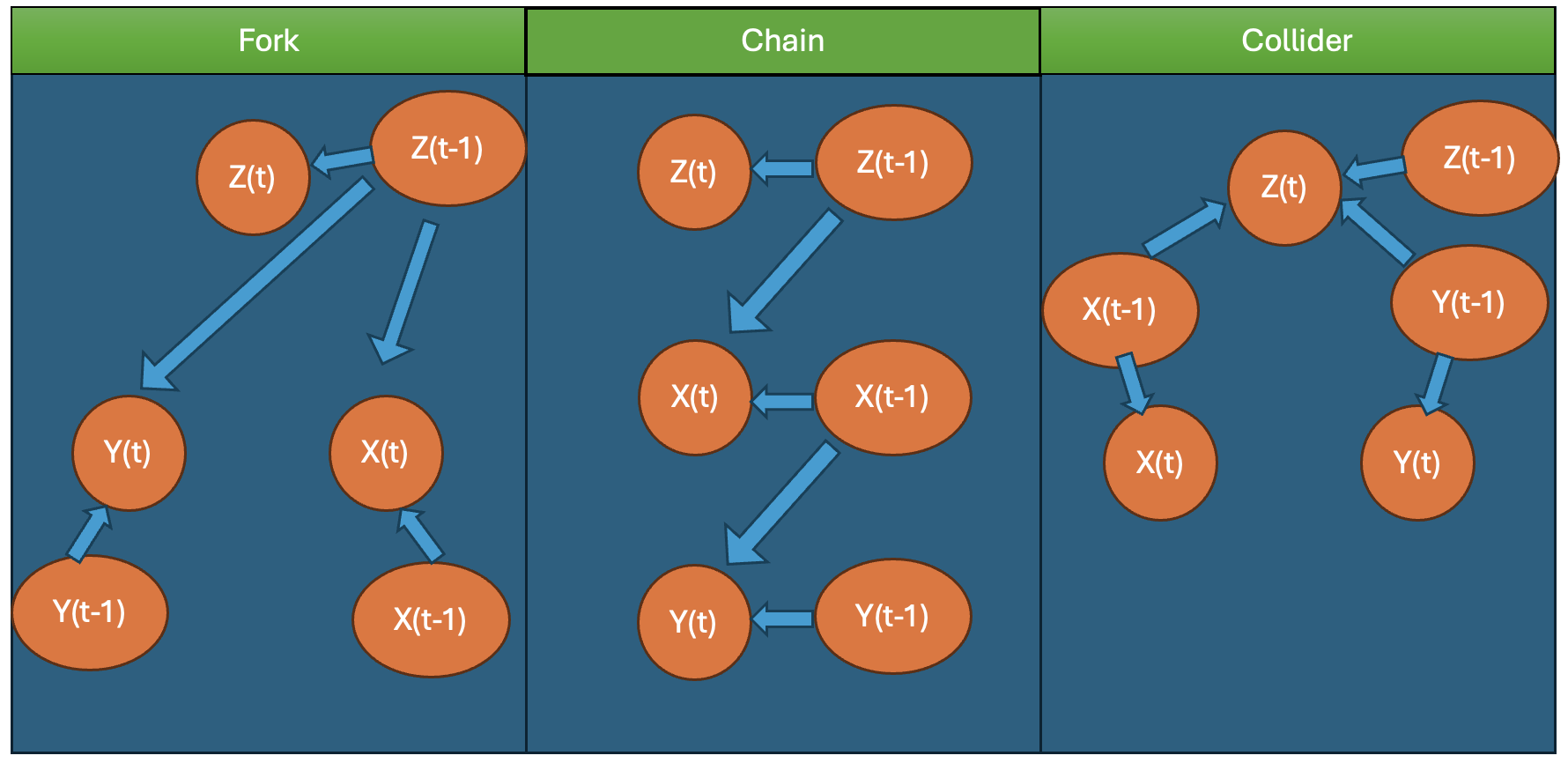}
    \caption{Chain, fork and collider structures visual representation. In the fork structure the past of $Z(t)$ granger causes the future value of  $X(t)$ and $Y(t)$. In the chain structure the past of $Z(t)$ granger causes $X(t)$ and the past of $X(t)$ granger causes $Y(t)$. In collider structure the past of $X(t)$ and $Y(t)$ granger causes $Z(t)$.}
    \label{fig:CFC}
\end{figure}

Chain fork and collider are the three main structures that appear in causal inference. Suppose X(t), Y(t), Z(t) are three real valued time series. We will refer to these as X, Y and Z. We can say that the Granger causal structure is that of a chain if the past of X is predictive of the future of Y, and the past of Y is predictive of the future of Z. Fork is the second causal structure which appears most frequently in causal inference literature. Fork arises when there is a common cause of two effects. In the absence of the cause the effects may appears to be related to each other. In terms of Granger causation in time series data, the fork structure will appear if the past of time series Z Granger causes the future of both time series X and Y. The third important structure is that of collider. Collider structure appears when there is a common effect of two separate causes. In terms of Granger causation for time series data it will appear when the past of time series X and Y Granger cause time series Z. To include self loops we can make the past of each time series predictive of its own future.
\par
Multivariate time series data is generated using autoregressive (AR) processes with linear and non-linear Granger causal components. Each time series $X_t(i)$ is modeled as an autoregressive process of order $p$, influenced by its own past and by the past of its parent nodes:

\[
X_t(i) = \sum_{k=1}^{K} \phi_k^{(i)} X_{t-k}(i) + \sum_{j \in \mathrm{Pa}(i)} f_{ij}(X_{t-1}(j)) + \varepsilon_t(i)
\]

Where $K$ denotes the autoregressive (AR) order, which is fixed to 2 in all our experiments. The autoregressive coefficients are given by \( \phi_k^{(i)} = 0.5^k \). For each variable \( i \), \( \mathrm{Pa}(i) \) denotes the set of its parent variables, and \( f_{ij}(\cdot) \) represents the influence function from variable \( j \) to \( i \). The noise term \( \varepsilon_t(i) \) is assumed to be zero-mean Gaussian noise with variance \( \sigma_e^2 \).

For nonlinear interactions, the influence function is defined as:
\[
f_{ij}(X_{t-1}(j)) = \tanh(|X_{t-1}(j)|) + \sin(|X_{t-1}(j)|)
\]

\begin{table}[ht]
\centering
%\renewcommand{\arraystretch}{1.2}
% --------- Chain ----------
\textbf{Chain Structure}
\vspace{0.1cm}

\textit{AUCROC}
\begin{tabular}{l|ccc}
\hline
\textbf{Training} & LSTM & CNN & Transformer \\
\hline
No ILD & 1.00 ± 0.00 & 1.00 ± 0.00 & 1.00 ± 0.00 \\
DP ILD & 0.90 ± 0.16 & 0.98 ± 0.05 & 0.95 ± 0.10 \\
ILD    & 0.60 ± 0.29 & 0.60 ± 0.29 & 0.61 ± 0.28 \\
\hline
\end{tabular}

\vspace{0.1cm}
\textit{AUPRC}
\begin{tabular}{l|ccc}
\hline
\textbf{Training} & LSTM & CNN & Transformer \\
\hline
No ILD & 1.00 ± 0.00 & 1.00 ± 0.00 & 1.00 ± 0.00 \\
DP ILD & 0.91 ± 0.14 & 0.97 ± 0.07 & 0.95 ± 0.10 \\
ILD    & 0.60 ± 0.28 & 0.60 ± 0.28 & 0.62 ± 0.26 \\
\hline
\end{tabular}

% --------- Fork ----------
\vspace{0.1cm}
\textbf{Fork Structure}

\vspace{0.1cm}
\textit{AUCROC}
\begin{tabular}{l|ccc}
\hline
\textbf{Training} & LSTM & CNN & Transformer \\
\hline
No ILD & 1.00 ± 0.00 & 1.00 ± 0.00 & 1.00 ± 0.00 \\
DP ILD & 1.00 ± 0.00 & 1.00 ± 0.00 & 1.00 ± 0.00 \\
ILD    & 0.86 ± 0.19 & 0.84 ± 0.24 & 0.89 ± 0.14 \\
\hline
\end{tabular}

\vspace{0.1cm}
\textit{AUPRC}
\begin{tabular}{l|ccc}
\hline
\textbf{Training} & LSTM & CNN & Transformer \\
\hline
No ILD & 1.00 ± 0.00 & 1.00 ± 0.00 & 1.00 ± 0.00 \\
DP ILD & 1.00 ± 0.00 & 1.00 ± 0.00 & 1.00 ± 0.00 \\
ILD    & 0.82 ± 0.26 & 0.80 ± 0.29 & 0.87 ± 0.17 \\
\hline
\end{tabular}

% --------- Collider ----------
\vspace{0.1cm}
\textbf{Collider Structure}

\vspace{0.1cm}
\textit{AUCROC}
\begin{tabular}{l|ccc}
\hline
\textbf{Training} & LSTM & CNN & Transformer \\
\hline
No ILD & 1.00 ± 0.00 & 1.00 ± 0.00 & 1.00 ± 0.00 \\
DP ILD & 0.90 ± 0.14 & 0.88 ± 0.18 & 1.00 ± 0.00 \\
ILD    & 0.58 ± 0.35 & 0.70 ± 0.34 & 0.71 ± 0.23 \\
\hline
\end{tabular}

\vspace{0.1cm}
\textit{AUPRC}
\begin{tabular}{l|ccc}
\hline
\textbf{Training} & LSTM & CNN & Transformer \\
\hline
No ILD & 1.00 ± 0.00 & 1.00 ± 0.00 & 1.00 ± 0.00 \\
DP ILD & 0.87 ± 0.19 & 0.86 ± 0.20 & 1.00 ± 0.00 \\
ILD    & 0.58 ± 0.34 & 0.67 ± 0.37 & 0.70 ± 0.25 \\
\hline
\end{tabular}
\caption{Performance comparison of LSTM, CNN, and Transformer models on Chain, Fork, and Collider structures under three training regimes: No ILD, Dual Pass ILD (DP ILD), and ILD. Values are mean $\pm$ standard deviation across 5 random datasets.}
\label{tab:all_structures_results}
\end{table}

Performance of different base encoder models in learning the Granger causality from 3 caual structures is given in table \ref{tab:all_structures_results}. Value of $\sigma^2_{e}$ was set to 0.1. We observe in our simulations that all of the models can learn the true Granger causal structure in the regular training regime where there is just dropout active in the hidden layers, which is referred to as No Input Layer Dropout (No ILD). Training neural network with dual passes with and without input layer dropout is referred as Dual Pass Input Layer Dropout (DP ILD). The last setting is the training with Input Layer Dropout (ILD). Hidden layers dropout is active in both ILD and DP ILD setting. Dropout in the input layer hinders the models ability to learn the true Granger causation. One possible explanation for this can be that in the absence of a true Granger causal time series the model may learn a mapping from correlated time series which may not be a true Granger cause. The Transformer model performs well across the three simulation and training regimes. 

\subsection{VAR}
\begin{figure}[t]
    \centering
    \includegraphics[width=1.0\linewidth]{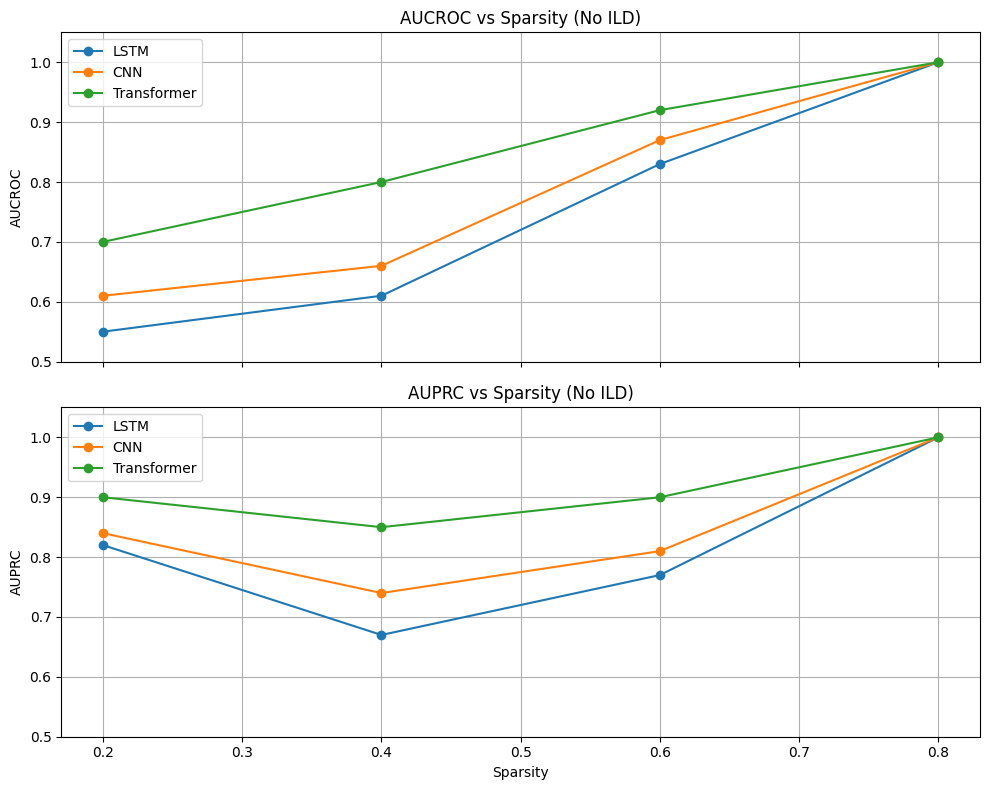}
    \caption{Comparison of Transformer, CNN and LSTM based on the mean AUROC and AUPRC across 5 random datasets, trained on different levels of sparsity for the VAR(2) data under No-ILD training.}
    \label{fig:Sparse_VAR}
\end{figure}

A P dimensional VAR(K) model can generate the data according to causal structure encoded in the mixing matrices $A_{k}$. A VAR(K) model generates data according to following linear dynamics
\begin{equation} 
    X(t) = \sum_{k = 1}^{K} A(k) X(t-k) + e(t)
\end{equation}
Where $e(t)$ is a Gaussian noise with mean zero and variance $\sigma^2_{e}$. We generate data with VAR(2) model with $T = 1000$, under different sparsity conditions. For more details for the simulation settings we refer the interested reader to \citep{tank2021neural}. Table \ref{tab:VAR_Comparison} lists the performance of different models under a sparse ground truth causal structure. We observe that un-restricted models LSTM, CNN and Transformer (No-ILD) excel at learning the true causal structure from the time series data where the task is just prediction. These models perform on-par with the standard sparse regression-based methods popular in neural Granger causal discovery. We also observe that ILD impacts the ability of these models to learn the Granger causal structure as the model may learn to encode spurious correlations in the hidden layers to excel at the task of prediction. DP-ILD appears to have caused less damage to the ability of these models to learn the ground truth Granger causal structure; however, it introduces an extra computational step of forward propagation with and without input layer dropout. Therefore in light of our simulation study, we do not recommend following this training procedure, as No-ILD can still recover the ground truth and is computationally efficient as well. Next we generate the data by varying the sparsity level. The comparison of the models under No-ILD is given in the figure \ref{fig:Sparse_VAR}. We observe that the Transformer model performs consistently well as compared to LSTM and CNN. As the sparsity level increases, all models perform equally well to discover the Granger causality.

\begin{table}[t]
\centering
\renewcommand{\arraystretch}{1.2}
\begin{tabular}{l|cc}
\hline
\textbf{Model - Regime} & \textbf{AUCROC} & \textbf{AUPRC} \\
\hline
LSTM - No ILD             & 1.00 ± 0.00 & 1.00 ± 0.00 \\
LSTM - DP ILD             & 0.98 ± 0.04 & 0.96 ± 0.05 \\
LSTM - ILD                & 0.80 ± 0.06 & 0.62 ± 0.04 \\
CNN - No ILD              & 1.00 ± 0.00 & 1.00 ± 0.00 \\
CNN - DP ILD              & 0.98 ± 0.03 & 0.92 ± 0.11 \\
CNN - ILD                 & 0.88 ± 0.10 & 0.70 ± 0.19 \\
Transformer - No ILD      & 1.00 ± 0.00          & 1.00 ± 0.00          \\
Transformer - DP ILD      & 1.00 ± 0.00         &    0.99 ± 0.01      \\
Transformer - ILD         & 0.70 ± 0.23         &   0.52 ± 0.36       \\
GVAR                      & 0.98 ± 0.01 & 0.92 ± 0.02 \\
CUTS                      &  1.00 ± 0.00        &   1.00 ± 0.00      \\
cLSTM                     & 0.98 ± 0.01         & 0.83 ± 0.07\\
\hline
\end{tabular}
\caption{VAR Performance Comparison (AUCROC and AUPRC). Values are mean $\pm$ standard deviation across 5 random datasets.}
\label{tab:VAR_Comparison}
\end{table}

\subsection{Lorenz-96}
Another benchmark simulation in the literature of neural GC is of Lorenz-96. The Lorenz 96 system in $p$ variables is defined as:
$$
\frac{dx_i}{dt} = (x_{i+1} - x_{i-2}) x_{i-1} - x_i + F, \quad \text{for} \quad 1 \leq i \leq p
$$
with periodic boundary conditions:
$$
x_{-1} = x_{p-1}, \quad x_0 = x_p, \quad x_{p+1} = x_1
$$
Here, \( F \) is a constant forcing term  that controls the degree of chaos in the system. The results are given in table \ref{tab: Lorenz_20_results}.The data were generated using $F = 20$, producing a 20-dimensional time series with $T = 1000 $ time points. The underlying ground truth structure is sparse. We find an interesting insight that unrestricted LSTM, CNN and Transformer model works very well at discovering the task of true Granger Causality from just prediction. These models perform at par with the standard sparse regression based component-wise deep learning models. This indicates that prediction-based joint modeling can also be a good baseline for comparison with more complex models that involve tuning of extra hyper-parameters; for instance, GVAR has a sparsity and smoothing parameters that have to be tuned using cross-validation. We observe that the performance of the models slightly degrades with input layer dropout. CNN here is observed to be mostly impacted in case of input layer dropout. One reason for this could be parameter sharing that is there in CNN architecture.

\begin{table}[ht]
\centering
\renewcommand{\arraystretch}{1.2}
\begin{tabular}{l|cc}
\hline
\textbf{Model - Regime} & \textbf{AUCROC} & \textbf{AUPRC} \\
\hline
LSTM - No ILD             & 1.00 ± 0.00 & 1.00 ± 0.00 \\
LSTM - DP ILD             & 1.00 ± 0.00 & 1.00 ± 0.01 \\
LSTM - ILD                & 1.00 ± 0.00         & 0.99 ± 0.01         \\
CNN - No ILD              & 0.98 ± 0.01 & 0.91 ± 0.03 \\
CNN - DP ILD              &    0.98 ± 0.01      &  0.93 ± 0.02        \\
CNN - ILD                 & 0.95 ± 0.01         & 0.88 ± 0.03         \\
Transformer - No ILD      & 1.00 ± 0.00 & 1.00 ± 0.00 \\
Transformer - DP ILD      & 1.00 ± 0.00 & 1.00 ± 0.00 \\
Transformer - ILD         & 1.00 ± 0.01 & 1.00 ± 0.01 \\
GVAR                      & 1.00 ± 0.00 & 1.00 ± 0.00 \\
CUTS                      & 0.96 ± 0.03         & 0.90 ± 0.02         \\
cLSTM                     &   0.93 ± 0.02      &  0.71 ± 0.04         \\
\hline
\end{tabular}
\caption{Lorenz-20 Performance Comparison (AUCROC and AUPRC). Values are mean $\pm$ standard deviation across 5 random datasets.}
\label{tab: Lorenz_20_results}
\end{table}

\subsection{Netsim}
Netsim  functional Magnetic Resonance Imaging (fMRI) Blood Oxygen level Dependent (BOLD) time series simulations \citet{smith2011network} consists of realistic time series dataset generated using fMRI forward model. In our simulation we consider 5 data replicates from the simulation number 3, these are the same replicates which were also used in \cite{marcinkevivcs2021interpretable}. This multi-variate time series data has 15 dimensions and $T = 200$ time points. Therefore this is a data scarce scenario. The ground truth GC is very sparse. The comparison of the models is given in \ref{tab:Netsim_Results}. We observe that sparse regression based methods work better compared to No-ILD Transformer, LSTM and CNN, however still these models act as a good starting baseline in recovering the Granger causal structure. Transformer models performs slightly better in terms of AUPRC as compared to LSTM and CNN.

\begin{table}[ht]
\centering
\renewcommand{\arraystretch}{1.2}
\begin{tabular}{l|cc}
\hline
\textbf{Model - Regime} & \textbf{AUCROC} & \textbf{AUPRC} \\
\hline
LSTM - No ILD           & 0.60 ± 0.07 & 0.17 ± 0.04 \\
CNN - No ILD            & 0.60 ± 0.07 & 0.14 ± 0.04 \\
Transformer - No ILD    & 0.60 ± 0.07 & 0.21 ± 0.08 \\
GVAR                    & 0.67 ± 0.06 & 0.26 ± 0.07 \\
CUTS                    & 0.72 ± 0.09         & 0.25 ± 0.08        \\
cLSTM                   & 0.62 ± 0.13        &  0.13 ± 0.04         \\
\hline
\end{tabular}
\caption{Netsim Performance Comparison (AUCROC and AUPRC). Values are mean $\pm$ standard deviation across 5 random datasets.}
\label{tab:Netsim_Results}
\end{table}
\subsection{Case Study : EEG Data Analysis}
We analyze the 3 minute ( 1-4 minute interval) resting state EEG data for a total of 30 subjects from Alzheimer's and healthy controls from a publicly available data set. EEG was recording with resting state eyes closed protocol from 19 channels from the scalp. More details about the study can be found in \cite{miltiadous2023dataset}. We used the preprocessed data which is publicly available along with the details of the pre-processing pipeline. Transformer model with  with 2 attention heads , 2 hidden layers with 128 neurons  each, with dropout was trained with No ILD for each subject separately. We modeled the data by incorporating the past 10 lags to predict one future time step for 19 channels.
\begin{figure}[h]
    \centering
    \includegraphics[width=1.0\linewidth]{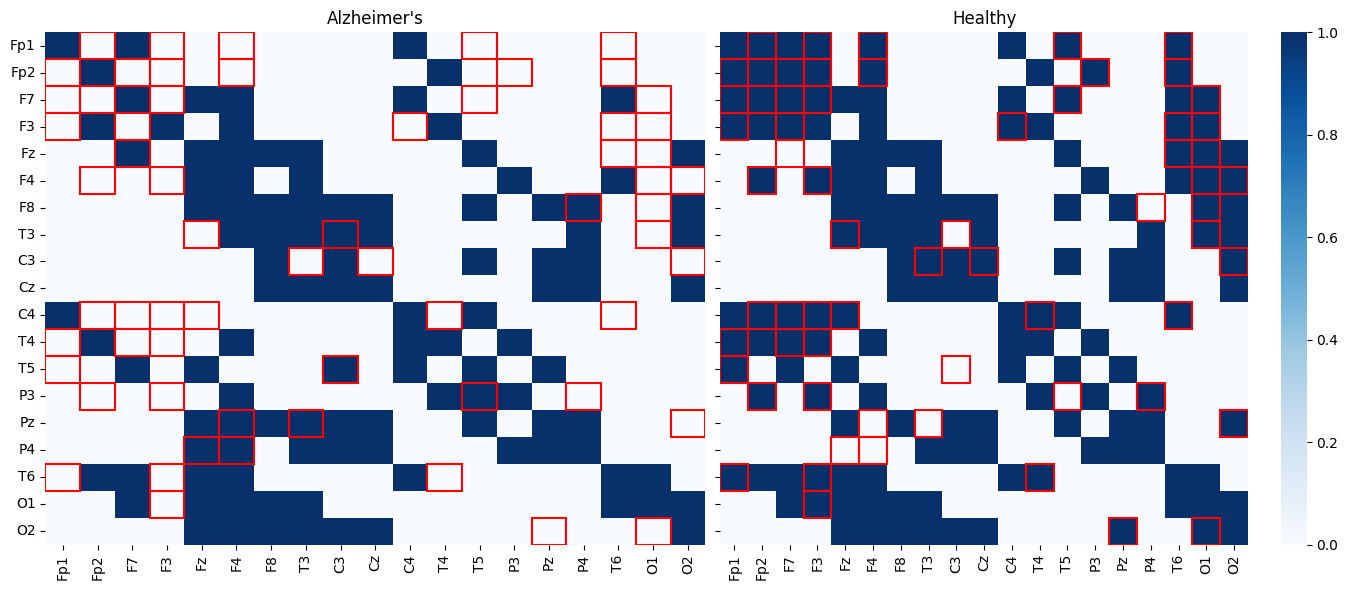}
    \caption{The mean Granger Causality between EEG channels for the Alzheimer's and Healthy controls. Blue depicts connectivity where are white depicts lack of a connection between granger causing channels in columns and granger caused channels in rows. Red square at any position indicates a different connectivity between two matrices.}
    \label{fig:EEG Analysis}
\end{figure}
\begin{figure}[h]
    \centering
    \includegraphics[width=0.5\linewidth]{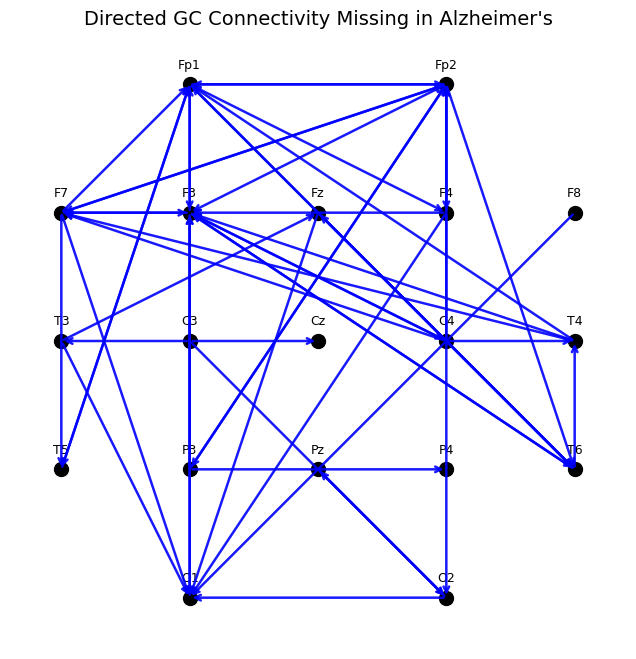}
    \caption{Directed GC present is the healthy controls and missing in the Alzheimer's group.}
    \label{fig:directed_GC}
\end{figure}

\begin{figure}[h]
    \centering
    \includegraphics[width=1.0\linewidth]{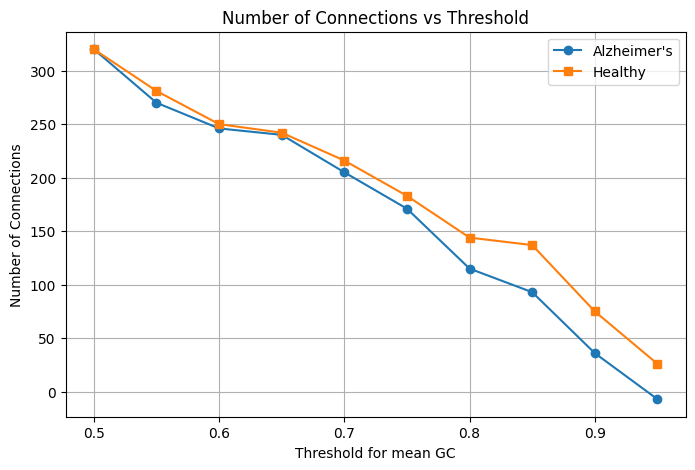}
    \caption{Number of connections as a function of threshold values. We observe that as the threshold becomes larger the difference between the mean GC of the two groups becomes more evident, where healthy controls show a strong connectivity.}
    \label{fig:GC_thresh}
\end{figure}
The mean connectivity matrices for both groups are shown in figure \ref{fig:EEG Analysis}. A separate model was trained for each subject in both groups, $GC_{s}$ for each subject was extracted from the respective model, average of the group GC was obtained by taking mean over all subjects in a group and the matrix was then threshold-ed at a value of $0.85$ to get a binary matrix. We observe from the figure \ref{fig:EEG Analysis} that the channels in the prefrontal lobe including $Fp1, Fp2, F7 $ and $F3$ are connected with each other densely where as it is missing in the Alzheimer's group. We also observe that for the healthy controls, $C4$ and $T4$ channels in the central and temporal lobe are granger caused by channels in the frontal, prefrontal lobe where as this behavior is missing in the Alzheimer's group. We also show that connectivity present in the healthy group but missing in the Alzheimer's group in figure \ref{fig:directed_GC}. We also plot the number of GC connections left in the mean connectivity matrix for each group after thresholding at a specific value. We can see that as we increase the threshold value the difference between the two groups become more and more evident, where the healthy controls show a strong connectivity as compared to the Alzheimer's group. Using default model network activity with fMRI \cite{greicius2004default} also reported a reduction in connectivity in the Alzheimer's group.
\section{Conclusion}
In this work we asked a simple question, can prediction alone reveal the Granger causal structure in time series data ?. We proposed a method involving dropout in the hidden layers to uncover the learned Granger causal structure from any pre-trained or trained time series model, by evaluating the distribution shift in the residuals of the predicted time series under general and reduced model forward propagation. It happens that under specific conditions a neural network which is being used to jointly model the time series data can infact learn the true Granger causal structure in the time series data. We also explore the setting where we drop few time series from the input data and predict the future evolution of all time series components. This is very similar to what is used in self supervised learning. We observe that in this specific setting the ability of the model to learn the true Granger causality degrades considerably as compared to un-restricted training or training without ILD. We also observe in almost all settings un-restricted training with joint modeling provides a good baseline for recovering the true Granger causality from the time series data, compared to sparse regression based methods that involve tuning of extra hyper-parameters.

\bibliographystyle{imsart-nameyear} % Style BST file
\bibliography{references}       % Bibliography file (usually '*.bib')

%% or include bibliography directly:
% \begin{thebibliography}{}
% \bibitem[\protect\citeauthoryear{???}{???}]{b1}
% \end{thebibliography}

\end{document}